\documentclass[sn-mathphys]{sn-jnl}

\usepackage{multirow}
\usepackage{amsmath}  
\usepackage{amssymb}  
\usepackage{color,array}
\usepackage{graphicx}
\usepackage{setspace}  
\newcommand{\tabincell}[2]{\begin{tabular}{@{}#1@{}}#2\end{tabular}}  
\usepackage{subfigure}  
\jyear{2021}%
\usepackage{booktabs}  
\usepackage{bm}  
\theoremstyle{thmstyleone}%
\theoremstyle{thmstyletwo}%

\theoremstyle{thmstylethree}%

\begin{document}

\title[Article Title]{Condition-Invariant and Compact Visual Place Description by Convolutional Autoencoder}


\author[1]{\fnm{Hanjing} \sur{Ye}}\email{teamedlar@gmail.com}

\author[2]{\fnm{Weinan} \sur{Chen}}\email{chenwn@sustech.edu.cn}

\author[2]{\fnm{Jingwen} \sur{Yu}}\email{11710414@mail.sustech.edu.cn}

\author[2]{\fnm{Li} \sur{He}}\email{hel@sustech.edu.cn}

\author[1]{\fnm{Yisheng} \sur{Guan}}\email{ysguan@gdut.edu.cn}

\author*[2]{\fnm{Hong} \sur{Zhang}}\email{hzhang@sustech.edu.cn}

\affil[1]{\orgdiv{Departmentof Mechanical and Electrical Engineering}, \orgname{Guangdong University of Technology}, \orgaddress{\city{Guangzhou}, \postcode{510006}, \country{China}}}

\affil[2]{\orgdiv{Department of Electrical and Electronic Engineering}, \orgname{Southern University of Science and Technology}, \orgaddress{\city{Shenzhen}, \postcode{518055}, \country{China}}}


\abstract{Visual place recognition (VPR) in condition-varying environments is still an open problem. Popular solutions are CNN-based image descriptors, which have been shown to outperform traditional image descriptors based on hand-crafted visual features. However, there are two drawbacks of current CNN-based descriptors: a) their high dimension and b) lack of generalization, leading to low efficiency and poor performance in applications. In this paper, we propose to use a convolutional autoencoder (CAE) to tackle this problem. We employ a high-level layer of a pre-trained CNN to generate features, and train a CAE to map the features to a low-dimensional space to improve the condition invariance property of the descriptor and reduce its dimension at the same time. We verify our method in three challenging datasets involving significant illumination changes, and our method is shown to be superior to the state-of-the-art. The code of our work is publicly available in \href{https://github.com/MedlarTea/CAE-VPR.}{https://github.com/MedlarTea/CAE-VPR}}

\keywords{Visual place recognition, Convolutional autoencoder, Unsupervised learning, Visual navigation}



\maketitle

\begin{figure}[t]
	\centering
	\includegraphics[width=0.9\textwidth]{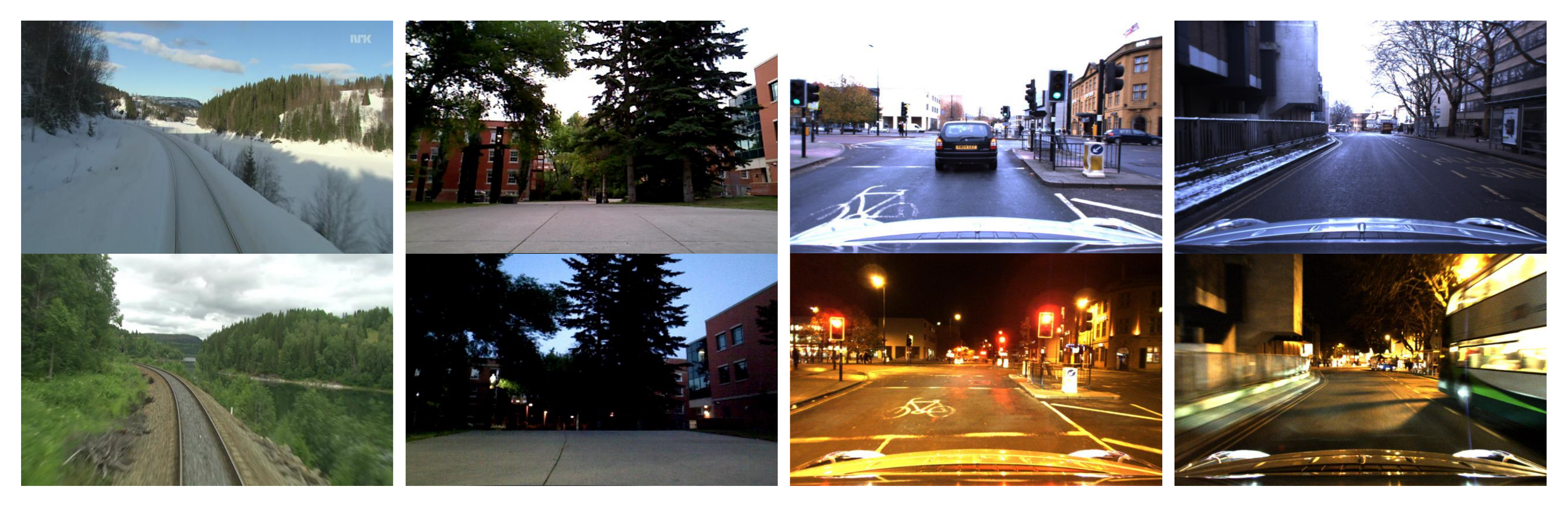}
	\caption{To improve visual place recognition, we employ a CAE to compress a CNN-generated descriptor and gain a condition-invariant and low-dimensional image descriptor. This figure has shown the effectiveness of our descriptor. The top row is query images (current robot view), and the below row is matching images that are successfully matched by our descriptor. From left to right, they are from summer-winter traverse of Norland \cite{olid2018single}, day-night scene of UACampus \cite{liu2015keypoint}, and autumn-night and snow-night sequences with dynamics of RobotCar \cite{maddern20171}.}
	\label{introduction}
\end{figure}

\section{INTRODUCTION}\label{sec1}

Visual place recognition (VPR) is essential in autonomous robot navigation.
VPR enables a robot to recognize previously visited places using visual data.
VPR provides loop closure information for a SLAM algorithm to obtain a globally consistent map.
Furthermore, VPR can support re-localization in a pre-built map of an environment.
Due to its essential role, many VPR methods \cite{lowry2015visual} have been proposed.
However, in long-term navigation tasks, significant appearance variation, typically caused by seasonal change, illumination change, weather change and dynamic objects, such as those shown in Fig. \ref{introduction}, is still a challenge to VPR.

VPR is typically formulated as an image matching procedure, which can be divided into two steps.
The first step of VPR, also known as loop closure detection in the literature, selects candidates where map images are represented by global descriptors and a matching procedure between the map images and the current robot view can be performed in terms of image similarity.
In the second step of VPR, verification is conducted via multi-view geometry, which uses keypoints in the images to determine if a query image (current robot view) is geometrically consistent with a candidate map image \cite{lowe2004distinctive, alahi2012freak, rublee2011orb, bay2008speeded}.
In this paper, we focus on the first step of loop closure detection, namely, generation of loop closure candidates efficiently and accurately.
Traditionally, a global descriptor is obtained by aggregating the handcrafted local descriptors, like SIFT \cite{lowe2004distinctive}, ORB \cite{rublee2011orb} and SURF \cite{bay2008speeded}.
In the case of significant appearance variations caused by, e.g., the day-night, season change and dynamic objects, handcrafted descriptors often fail to recognize places since locally keypoint descriptors can change significantly with the condition-dependent appearance.
Convolutional neural networks (CNNs) have shown their advantages in various visual recognition tasks \cite{krizhevsky2012imagenet, girshick2014rich, ronneberger2015u}  and have been used to generate global image descriptors for visual loop closure detection.
In \cite{sunderhauf2015performance}, a pre-trained CNN is firstly used to produce a global descriptor directly.
Alternatively, end-to-end trained descriptors with aggregating methods \cite{arandjelovic2016NetVLAD, radenovic2018fine, gordo2016deep} are proposed to gain higher performance. 

However, deep learning-based VPR methods have some limitations.
Firstly, a pre-trained CNN may generate descriptors easily with a dimension in the 10's of thousands, and hence result in time and storage problems.
Secondly, the generalization ability of CNN descriptors is often poor.
To tackle these limitations, we use a CAE to compress a CNN-generated descriptor and improve their ability to generalize.
Experiments on challenging datasets show that, by compressing the local feature maps of a CNN by CAE, the compressed descriptor achieves better results than the uncompressed descriptor in both seen and unseen environments at a lower computational cost.

\section{RELATED WORK}

\subsection{Visual Place Recognition}
In this paper, our concern is using a global descriptor to represent an image for loop closure detection. In the early works, VPR has been attained by extracting handcrafted local keypoints and descriptors firstly, such as SIFT \cite{lowe2004distinctive}, ORB \cite{rublee2011orb} and SURF \cite{bay2008speeded}. Then, these local features are aggregated to a global descriptor by vector quantization such as bag-of-words \cite{sivic2003video, jegou2010aggregating, philbin2007object}, VLAD \cite{arandjelovic2013all} and Fisher Vectors \cite{jegou2011aggregating}. Through clustering, a low-dimensional global descriptor can be achieved although spatial relations between the local descriptors are not encoded.
Although these traditional methods have been widely used in SLAM (simultaneous localization and mapping) research, they still struggle in large-scale environments with severe appearance changes \cite{lowry2015visual}.

Recently, researchers have proposed to use CNNs to extract features for loop closure detection in large-scale environments. At first, pre-trained classification CNNs are directly used to extract dense local feature maps \cite{sunderhauf2015performance,sharif2014cnn,babenko2014neural}, which serve as the visual features for visual place description. However, due to their high dimensions and inability to adapt to crowded environments, an end-to-end training model with a feature extractor and a pooling layer has been proposed, e.g., NetVLAD\cite{arandjelovic2016NetVLAD}, generalized-mean pooling \cite{radenovic2018fine}, max pooling \cite{tolias2015particular} and average pooling \cite{razavian2016visual}. Although end-to-end models can perform well in crowded environments with low dimensions, training bias is introduced by training datasets. It leads to a poor generalization of the end-to-end trained descriptors to unseen environments. Here, we use the unsupervised method of CAE to learn an image descriptor by minimizing the reconstruction loss of the high-level features of a CNN instead. This enables the encoded descriptor to attain discriminative features and generalize to unseen environments with a lower dimension.

\subsection{Convolutional Autoencoder}
CAE has shown its superior performance in many applications.
CGAN (conditional generative adversarial nets) \cite{mirza2014conditional} and pix2pix \cite{isola2017image} use CAE as their basic architecture to encode features and generate images from input images in a source domain to that in a target domain, such as from day to night, from labels to facade and from edges to a photo. Moreover, in U-Net \cite{ronneberger2015u}, semantic segmentation is achieved by a CAE-like architecture.

Recently, Madhu \emph{et al.} \cite{vankadari2020unsupervised} propose a CAE-based GAN to estimate depth maps from night-time images. It is worth noting that it also uses the descriptor from the encoder to accomplish the day-night VPR task.
Merrill \emph{et al.} \cite{merrill2018lightweight} utilize CAE to force the output of the decoder to be similar to the histogram of oriented gradients (HOG), and the output of the encoder is used as a global descriptor in the inference procedure.
Dai \cite{dai2020keypoint} uses a CAE to compress and fuse the local feature maps of the image patches for improving loop closure verification. Similar to Dai \cite{dai2020keypoint}, our method trains a CAE to reconstruct the local feature maps of the CNNs and then uses the resulting encoder for generating image descriptors. Differently from \cite{dai2020keypoint}, the CAE in our method reconstructs the feature maps of the whole image, instead of feature maps of local image patches. In this way, our encoder can capture the most relevant features of the whole image for VPR.

\begin{figure*}[t]
	\centering
	\includegraphics[width=\textwidth]{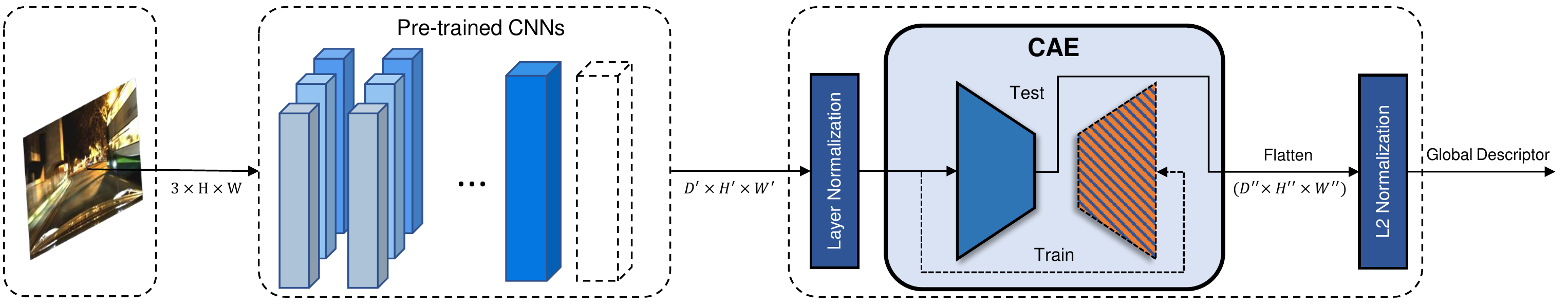}
	\caption{The detailed pipeline of our system. Given an image with $3\times{H}\times{W}$, CNNs  extract the local feature map $X_i$ with $D^{'}\times{H^{'}}\times{W^{'}}$. The CNNs are classification pre-trained or VPR-trained, e.g. AlexNet, VGG16. Both are cut at the last convolutional layer (conv5), before ReLU. In the training time, CAE is trained unsupervised by a reconstruction loss. In the test time, the decoder part of CAE is not involved and the encoder part is kept to compress the normalized feature map and produce a low-dimensional global descriptor with $D^{''}\times{H^{''}}\times{W^{''}}$. The global descriptor is then flattened and L2 normalized.}
	\label{pic_mehtod}
\end{figure*}

\section{APPROACH}

In this section, we describe our network architecture and training strategy.
The overall structure is shown in Fig. \ref{pic_mehtod}.
In our framework, a local feature map is extracted from a pre-trained CNN.
Specifically, the local feature map is extracted by a high-level layer of a pre-trained CNN.
The map is then normalized \cite{ba2016layer} and fed into the CAE.
In the training procedure, the CAE consisting of an encoder and a decoder is trained by a reconstruction loss. 
However, in the inference step, the decoder part is dropped and only the encoder part is kept to produce the image descriptor.

\subsection{Feature Extraction} \label{3-A}
Different layers of CNNs describe an image at different levels of semantics \cite{sunderhauf2015performance, hou2015convolutional}.
In the VPR task, we choose the feature map of a deep layer, which is found in previous works to be condition-invariant and low-dimensional.

Similar to \cite{arandjelovic2016NetVLAD}, we choose AlexNet \cite{krizhevsky2012imagenet} and VGG16 \cite{simonyan2014very} as our backbone. The local feature map $F$ is computed as:
\begin{equation}
	F = f_{\theta}(I)
\end{equation}
where $I$ is an input image with a dimension of $3\times{H}\times{W}$. $H$ and $W$ are the height and width of the input image. $f_{\theta}$ is a VPR-trained or pre-trained CNN without fine-tuning. In our work, $F$ is from the last convolution layer of a CNN, before ReLU. For AlexNet, the dimension of $F$ is $256\times{(\frac{1}{16}H-2)}\times{(\frac{1}{16}W-2)}$. For VGG16, the dimension is $512\times{\frac{1}{16}H}\times{\frac{1}{16}W}$. At such high dimensions, the global descriptor is of low computational efficiency to be stored and compared algorithmically for real-time performance. To tackle these problems, we use a CAE to compress the descriptor into a low-dimensional representation while promoting its condition-invariant capacity.

\begin{figure}[t]
	\centering
	\includegraphics[width=0.8\textwidth]{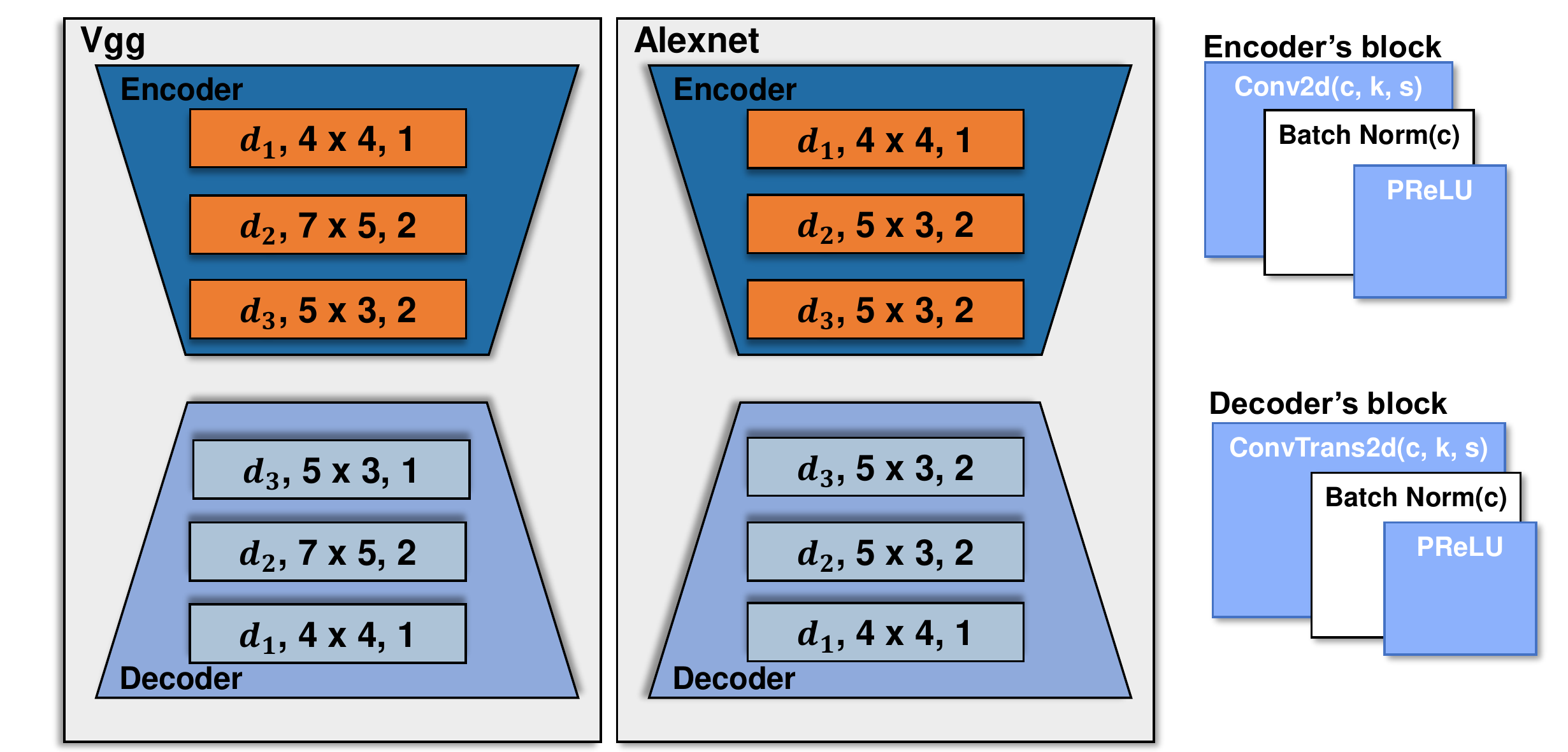}
	\caption{The architecture of our CAE. Every encoder block contains Conv2d, Batch Normalization and PReLU layer. In the decoder block, Conv2d is replaced with ConvTranspose2d. \textbf{c, k, s} in the picture means \textbf{channels, kernel size and stride}, respectively. \textbf{$d_3$} is a \textbf{changeable parameter} of the last convolutional kernel channels of the encoder, to produce the global descriptor with variable dimensions.}
	\label{CAE}
\end{figure}

\subsection{Convolutional Autoencoder} \label{3-B}
Given a high-dimensional local feature map $F$, we firstly normalize it with layer normalization \cite{ba2016layer}. Then, a CAE with three encoder layers and three decoder layers is trained to reconstruct the normalized feature map. The architecture and the whole training strategy are:
\begin{equation}
	\begin{aligned}
		&\hat{y} = g_{\theta}(h(F)) \\
		&g_{\theta} = \left [ g_{enc}\ g_{dec} \right ] \\
		&min\ (h(F),\ \hat{y})
	\end{aligned}
\end{equation}
where $h(x)$ is a layer normalization function \cite{ba2016layer}, $g_\theta$ is a CAE with an encoder $g_{enc}$ and a decoder $g_{dec}$. In the training procedure, we reconstruct the normalized local feature map $h(F)$ to train the CAE. We use mean squared error and back propagation to reconstruct the normalized feature map $h(F)$. The mean squared error is defined as:
\begin{equation}
	L_{mse} = \frac{1}{n}\left \| h(F)-g_{\theta}(h(F)) \right \|^2_2
\end{equation} 
where $n$ is the dimension of the local feature map $F$. The layer normalization $h(x)$ is defined as:
\begin{equation}
	h(x)=\frac{x-E[x]}{\sqrt{Var(x)+\epsilon}}
\end{equation}
where $x$ is a sample, $E[x]$ and $Var(x)$ are the mean and variance of the sample respectively, both of which are updated during training but frozen in the inference step. $\epsilon$ is a given value added to the denominator for numerical stability, which is set to $10^{-5}$  in our study.

Classic dimension reduction methods, e.g., PCA, only detect the linear relationship between features. For deep-learning-based pooling approaches, e.g., GeM \cite{radenovic2018fine}, max pooling \cite{tolias2015particular} and average pooling \cite{razavian2016visual}, they directly aggregate the $D^{'}\times{H^{'}\times{W^{'}}}$ local feature map into a descriptor with $D^{'}$ dimensions. Because the features across spatial dimensions are directly aggregated, the spatial information in the feature map is therefore lost. In contrast, our CAE compresses the feature map non-linearly while maintaining the spatial relationship.
In addition, the local feature map $F$ is usually sparse and high-dimensional \cite{chenonly2017}, indicating that only a few regions of a feature map have a solid response to a particular task like VPR or classification. With these attributes, our CAE can keep the most relevant features by reconstructing the input.

As shown in Fig. \ref{CAE}, in our CAE, each block in the encoder/decoder is composed of a convolutional/deconvolutional unit, a batch normalization unit \cite{ioffe2015batch} and a parametric rectified linear unit \cite{he2015delving}.

Since our CAE is based-on VGG16, the kernel sizes of three encoder blocks are $4\times{4},\ 7\times{5},\ 5\times{3}$, with strides 1, 2, 2, respectively. The channels of the first two encoder blocks are respectively $d_1$ and $d_2$. Similar to \cite{dai2020keypoint}, to generate descriptors of different dimensions for comparison, the channels of the last encoder block $d_3$ are accordingly set to 8, 16, 32, 64, 128, 256 and 512. For AlexNet, the kernel sizes of three encoder blocks are $4\times{4},\ 5\times{3},\ 5\times{3}$, and the strides are 1, 2, 2, respectively.
We adopt the same configurations as the encoder channels of VGG16 in our AlexNet encoder. For both architectures, the parameters of the decoder are similar to the encoder.

In the inference step, the decoder is not involved and the encoder is used to infer the compressed descriptor:
\begin{equation}
	X = g_{enc}(h(F))
\end{equation}
where $X$ is then flattened and L2-normalized to generate the final global descriptor.


\begin{table*}[t]
	\caption{Summary of the experimental datasets. \textbf{RobotCar (dbNight vs. qAutumn)} indicates that a night sequence is used as the reference set (database) and a autumn sequence is the query set.}
	\vspace{5pt}
	\centering
	\scalebox{0.6}{
	\begin{tabular}{lllll}
		\toprule
		\multirow{2}*{\textbf{Dataset}}  & \multirow{2}*{\textbf{Environment}} & \multicolumn{2}{c}{\textbf{Traverse}} & \multirow{2}*{\textbf{Appearance Change}} \\ \cline{3-4}
		&               &\textbf{Reference}        &\textbf{Query}                & \\
		\midrule  
		Nordland	            &Train Journey	&1415 (summer)	  &1415 (winter)	    &Very Strong\\
		UACampus	            &Campus	        &647 (night)	      &647 (day)	            &Very Strong\\
		
		RobotCar (dbNight vs. qAutumn) &Urban	        &7504 (night)	  &1046 (autumn)	        &Very Strong\\
		RobotCar (dbNight vs. qSnow)	  &Urban	        &7504 (night)	  &1043 (snow)	        &Very Strong\\
		RobotCar (dbSunCloud vs. qSnow)	    &Urban	        &7504 (sunCloud)	  &1043 (snow)	        &Strong\\
		RobotCar (dbSunCloud vs. qAutumn)	&Urban	        &7611 (sunCloud)	  &1046 (autumn)	        &Moderate\\
		\bottomrule
	\end{tabular}}
	\label{dataset}
\end{table*}

\section{EXPERIMENTS SETUP}

\subsection{Dataset}
To evaluate the performance of our proposed method, we use three datasets in the experiments. These datasets contain significant appearance changes, in urban, train track and university campus environments. Sample images from the datasets are shown in Fig. \ref{introduction}. The detailed description of the datasets is provided in Table \ref{dataset} as well as below.

\begin{figure}[t]
	\centering
	\includegraphics[width=0.7\textwidth]{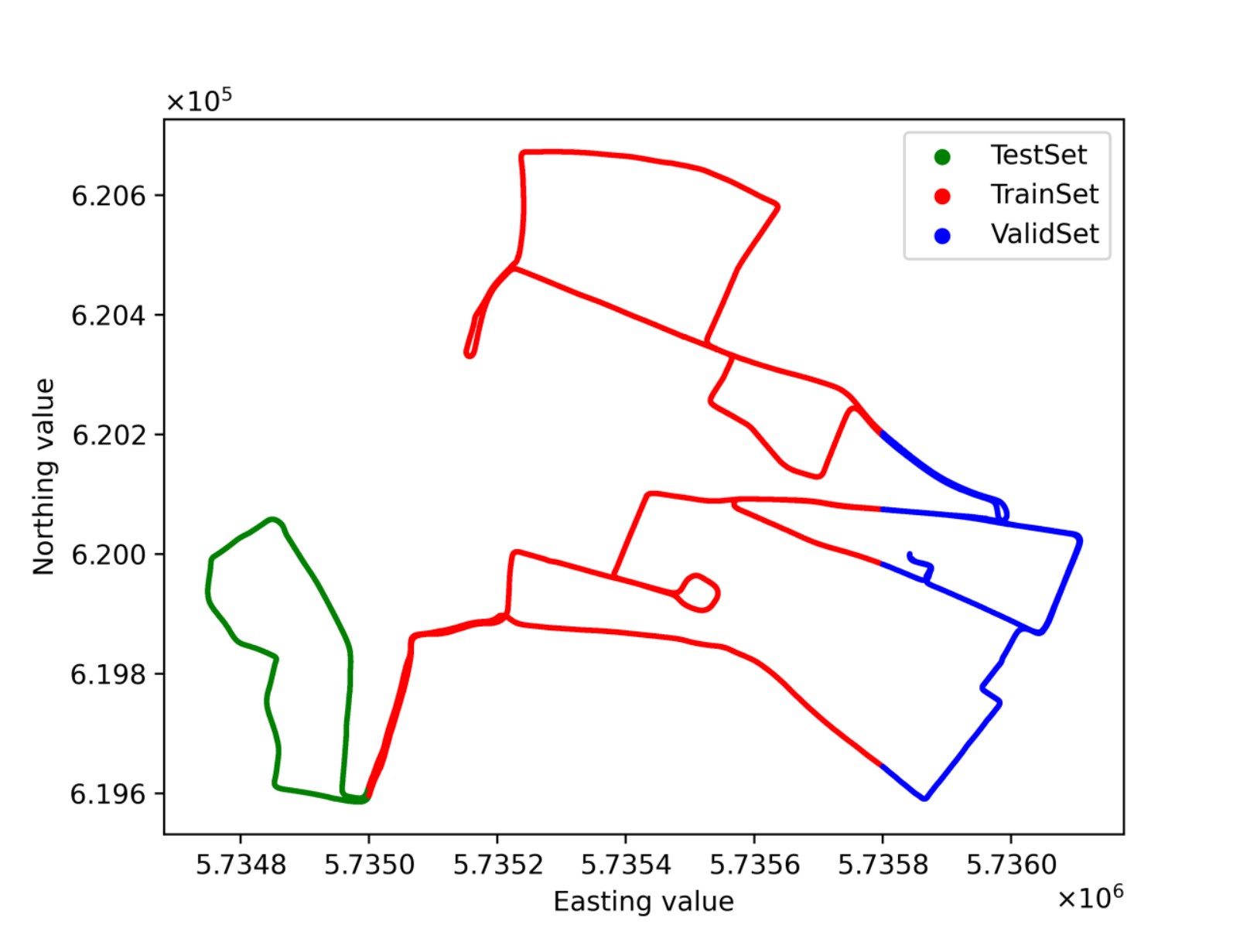}
	\caption{Dataset separation for RobotCar which is strictly geometrically non-overlapped. The red route is for CAE training, the green route represents the test set, and the route with blue indicates the validation set.}
	\label{robotcar}
\end{figure}

\noindent\textbf{Oxford RobotCar \cite{maddern20171}} is an urban dataset that records a 10km route through central Oxford multiple times over one year. Within this dataset, challenging views with appearance changes are captured due to season, weather and time of the day. We choose a subset consisting of five sequences$\footnote{suncloud: 2014-12-09-13-21-02, night: 2014-12-10-18-10-50, 2014-12-16-18-44-24, autumn: 2014-11-18-13-20-12, snow: 2015-02-03-08-45-10}$ involving sun cloud, autumn, snow and night environments, all of which contain strong appearance changes.
To validate the effectiveness of our method, the dataset is separated with no overlap. Specifically, we extract a front-view image per meter for all sequences to construct the datasets.
As shown in Fig. \ref{robotcar}, the red route is the training set which includes 24k images, the green route is the test set, and the blue route represents the validation set.
In the matching procedure, we have a query and a reference set. The query set contains the images of the green route, and the reference set includes the images of the whole route to increase the difficulty of matching.
If the distance between a matched pair is within 25m, the decision is considered as a true positive.

\noindent\textbf{Nordland \cite{olid2018single}} is a train journey dataset that contains significant seasonal changes. In this paper, summer and winter traverse are used as reference and query respectively. If the reference image is within two frames relative to the query, it is treated as a true positive.

\noindent\textbf{UACampus \cite{liu2015keypoint}} is a campus dataset with day-night illumination changes recorded on campus of University of Alberta. Here, two subsets were captured in the morning (06:20) and evening (22:15) along the same route. The ground truth matching is available by manual annotation.

\subsection{Evaluation Metric}
\noindent\textbf{Recall@1,5,10.} To verify the overall performance of an image descriptor, we follow the common evaluation metric defined in  \cite{arandjelovic2016NetVLAD}, which is based on the top $K$ nearest neighbors among all database descriptors to a query one. Matching is considered successful if the correct match exists within the top $K$ nearest pairs. $K$ is set to 1, 5 and 10 in our experiments.

\noindent\textbf{Precision-Recall Curve.} Precision-Recall is another key evaluation metric in VPR. Given matched pairs and a threshold in terms of cosine similarity between image descriptors, we have the numbers of true positives, false positives and false negatives. precision and recall are defined as:
\begin{equation}
	\begin{aligned}
		Precision &= \frac{TP}{TP+FP}\\
		Recall &= \frac{TP}{TP+FN}
	\end{aligned}
\end{equation} 
Multiple pairs of precision-recall (PR) values are produced by varying the threshold, and a PR-curve can then be plotted for evaluation. A high threshold often causes low recall and high precision because a strict matching policy always reduces false positives (FP) but at the cost of many false negatives (FN). The ideal performance is when both precision and recall are high.

\noindent\textbf{Average Precision (AP).} The overall performance is usually represented by the average precision. It summarizes a precision-recall curve as the weighted mean of precisions achieved at all recall values:
\begin{equation}
	AP = \sum_{n}(R_n-R_{n-1})P_n
\end{equation}
where $P_n$ and $R_n$ are the precision and recall, respectively. $n$ is the $n_{th}$ threshold. Intuitively, AP is the integral of the precision-recall curve.

\noindent\textbf{L2 distance distribution.} To test the discriminative capacity of our global descriptor, we draw the histogram distribution of L2 distance of the true matches and the false matches.
Their mean values are calculated to quantify the distinguishable ability of the descriptor. Intuitively, a small overlap of two distributions indicates good discrimination.

\subsection{Baseline and Our Method}

\textbf{NetVLAD} \cite{arandjelovic2016NetVLAD} is a popular method trained on Pitts30k. We choose the best model by evaluating the method on the Pitts30k-valid. After training, we compress the descriptor dimension to 4096 with PCA and whitening. In our experiments, we use four tuples (one query, one positive, one negative) for training, for the purpose of reducing computational resources usage.
\noindent\textbf{VGG16} \cite{simonyan2014very} is the backbone of the Pitts30k-trained NetVLAD, and outputs the descriptor with $512\times{(\frac{1}{16}H)}\times{(\frac{1}{16}H)}$ dimensions.
\noindent\textbf{AlexNet} \cite{krizhevsky2012imagenet} is of the matconvnet version pre-trained in ImageNet, with the descriptor of  $256\times{(\frac{1}{16}H-2)}\times{(\frac{1}{16}H-2)}$ dimensions.
\noindent\textbf{OursV} is composed of \textbf{VGG16} and the CAE introduced in Section \ref{3-B}. 
\noindent\textbf{OursA} consists of \textbf{AlexNet} and our CAE.

\begin{table}[ht]
	\caption{Comparison of different settings of spatial dimensions and channel dimensions of CAE in terms of R@1. \textbf{$d_1$}, \textbf{$d_2$} and \textbf{$d_3$} represent the number of channels of the three encoder blocks. \textit{c1}  and \textit{c2} are different output spatial dimensions of the encoder module. This experiment is conducted on RobotCar (dbNight vs. qSnow).}
	\vspace{5pt}
	\centering
	\scalebox{0.9}{
		\begin{tabular}{cccccccc}
			\toprule
			\textbf{Method} &\bm{$d_1$} &\bm{$d_2$} &\bm{$d_3$} &\textbf{\textit{c1}} &\textbf{\textit{c2}} &\textbf{Output Dimension} &\textbf{R@1}\\ \hline
			oursA &128    &128	&256 &\checkmark &$\times$ &8192 &0.739 \\
			&256  &256	    &256 &\checkmark &$\times$ &8192 &0.742 \\ 
			&512  &256	    &256 &\checkmark &$\times$ &8192 &0.755 \\ 
			&512  &512	    &256 &\checkmark &$\times$ &8192 &0.734 \\ 
			&512  &1024	&256 &\checkmark &$\times$ &8192 &0.738 \\ \cline{2-8}
			&256  &256	    &12 &$\times$ &\checkmark &8448 &0.549 \\ 
			&512  &256	    &12 &$\times$ &\checkmark &8448 &0.559 \\ 
			&512  &512	    &12 &$\times$ &\checkmark &8448 &0.508 \\ 
			&512  &1024	&12 &$\times$ &\checkmark &8448 &0.527 \\ \midrule
			
			oursV  &128     &128	&256 &\checkmark &$\times$ &8192 &0.860 \\
			&256  &256	    &256 &\checkmark &$\times$ &8192 &0.869 \\ 
			&512  &256	    &256 &\checkmark &$\times$ &8192 &0.872 \\ 
			&512  &512	    &256 &\checkmark &$\times$ &8192 &0.879 \\ 
			&512  &1024	&256 &\checkmark &$\times$ &8192 &0.879 \\ \cline{2-8}
			&256  &256	    &10 &$\times$ &\checkmark &8160 &0.806 \\ 
			&512  &256	    &10 &$\times$ &\checkmark &8160 &0.804 \\ 
			&512  &512	    &10 &$\times$ &\checkmark &8160 &0.809 \\ 
			&512  &1024	&10 &$\times$ &\checkmark &8160 &0.832 \\
			\bottomrule
	\end{tabular}}
	\label{ablation}
\end{table}

\subsection{Implementation Details}
In the experiments, the resolution of the input image is $640\times{480}$. For VGG16, the local feature map has a dimension of 614,400. For AlexNet, the dimension is 272,384. The hyperparameters of our CAE are optimized empirically by experiments conducted in RobotCar (dbNight vs. qSnow). The results are shown in Table \ref{ablation} where $d_1$, $d_2$ and $d_3$ represent the number of channels of the three encoder blocks, \textit{c1} and \textit{c2} imply the output spatial dimensions. Specifically, \textit{c1} adopts the original settings and \textit{c2} has a size of $3\times{3}$ and the stride is $1$. For a balance of effectiveness and computation resource, $d_1$ and $d_2$ are set to 128, and \textit{c1} is adopted.

During the CAE training, the backbone CNN is frozen. The Adam optimization algorithm is used to learn the model parameters, with a learning rate of 0.001 and a batch size of 128. The model is trained for 50 epochs. All the training is executed in PyTorch with 4 TITAN XP.

\begin{table}[ht]
	\caption{Comparison of the baselines and our methods in terms of average precision (AP) and recall@1, 5, 10.}
	\vspace{5pt}
	\centering
	\scalebox{0.7}{
		\begin{tabular}{cccccc}
			\toprule
			\textbf{Dataset}  &\textbf{Method} &\textbf{AP} &\textbf{R@1} &\textbf{R@5} &\textbf{R@10}\\
			\midrule
			\multirow{5}*{Nordland}	    &NetVLAD	&0.402  &0.273	&0.473 &0.576\\ 
			&AlexNet	    &0.956  &0.910	&0.976 &0.992\\ 
			&VGG16	        &0.815  &0.634	&0.819 &0.864\\ 
			&OursA (4096d)	&0.984 &\textbf{0.962} &\textbf{0.996} &\textbf{0.999}\\
			&OursV (4096d)	&\textbf{0.979} &0.946 &0.990 &0.997\\
			\midrule
			
			\multirow{5}*{UACampus}	        &NetVLAD	&0.744  &0.674	&0.788 &0.838\\ 
			&AlexNet	    &0.993  &0.932	&0.974 &0.985\\ 
			&VGG16	        &0.996  &0.934	&0.971 &0.986\\ 
			&OursA (4096d)	&\textbf{0.999} &\textbf{0.977} &\textbf{0.994} &\textbf{0.995}\\ 
			&OursV (4096d)	&0.999 &0.969 &0.992 &\textbf{0.995}\\
			\midrule
			
			\multirow{5}*{\tabincell{c}{RobotCar\\ (dbNight vs. qAutumn)}}	   &NetVLAD	&0.933  &0.759	&0.874 &0.914\\ 
			&AlexNet	&0.950 &0.827 &0.879 &0.906\\ 
			&VGG16	&0.865  &0.644	&0.719 &0.754\\ 
			&OursA (4096d)	&0.952 &0.832 &0.884 &0.903\\ 
			&OursV (4096d)	&\textbf{0.987} &\textbf{0.881} &\textbf{0.913} &\textbf{0.928}\\
			\midrule
			
			\multirow{5}*{\tabincell{c}{RobotCar\\ (dbNight vs. qSnow)}}	   &NetVLAD	&0.893  &0.691	&0.816 &0.849\\ 
			&AlexNet	&0.657  &0.453	&0.541 &0.593\\ 
			&VGG16	&0.810  &0.523	&0.584 &0.635\\ 
			&OursA (4096d)	&0.950 &0.730 &0.797 &0.825\\ 
			&OursV (4096d)	&\textbf{0.975} &\textbf{0.861} &\textbf{0.891} &\textbf{0.907}\\
			\midrule
			
			\multirow{5}*{\tabincell{c}{RobotCar\\ (dbSunCloud vs. qSnow)}}	   &NetVLAD	&0.991  &0.877	&0.911 &0.934\\ 
			&AlexNet	&0.946  &0.803	&0.899 &0.916\\ 
			&VGG16	&0.943  &0.776	&0.830 &0.860\\ 
			&OursA (4096d)	&0.989 &0.868 &0.932 &0.947\\ 
			&OursV (4096d)	&\textbf{0.995} &\textbf{0.919} &\textbf{0.941} &\textbf{0.952}\\
			\midrule
			
			\multirow{5}*{\tabincell{c}{RobotCar\\ (dbSunCloud vs. qAutumn)}} &NetVLAD	&0.996  &0.928	&0.956 &0.965\\ 
			&AlexNet	&0.991 &0.902 &0.932 &0.942\\ 
			&VGG16	&0.972  &0.820	&0.879 &0.906\\ 
			&OursA (4096d)	&0.992 &0.909 &0.931 &0.945\\ 
			&OursV (4096d)	&\textbf{0.996} &\textbf{0.930} &\textbf{0.962} &\textbf{0.971}\\ 
			\bottomrule
			
	\end{tabular}}
	\label{recall}
\end{table}

\section{RESULTS AND DISCUSSION}

\subsection{Effectiveness and Stability}

We first compare the performance of our method representative CNN-based image descriptors, namely, NetVLAD and those from VGG16 and AlexNet. In Table \ref{recall}, we can observe that NetVLAD performs better on RobotCar than on Norland and UACampus. VGG16 (the backbone of NetVlad) shows quite different results in this test. On Norland, VGG16 surpasses NetVLAD by a significant margin with an AP of 0.815 versus 0.402 and recall@1 of 0.634 versus 0.273. Nevertheless, it is slightly worse with an AP of 0.865 versus 0.933 on RobotCar (dbNight vs. qAutumn). This result could be attributed to the training bias introduced by the Pitts30k dataset, which is also an urban dataset similar to RobotCar. For VGG16, only conv5 and the following layer are trained. NetVLAD, with VGG16 as its backbone, includes a deep-learning-based VLAD module. Furthermore, the deep-learning-based VLAD module is optimized in the clustering space of the training datasets.

Compared to NetVLAD and VGG16, OursV achieves better results with a higher AP, with the output dimension set to 4096 for a fair comparison. Even on a dataset with large appearance changes, such as RobotCar (dbNight vs. qSnow), the recall@1 of OursV is 0.861 versus NetVLAD's 0.691 and VGG16's 0.523. It is worth noting that our method is unsupervised in this experiment on RobotCar, and it can nonetheless perform well in the non-urban dataset like Norland and UACampus.
To further validate the effectiveness and generalization ability of our method, we conduct experiments with different feature extractors, such as AlexNet pre-trained on ImageNet. OursA, which is also of dimension 4096, always produces better results than AlexNet, with an AP of 0.975 versus 0.657 in RobotCar (dbNight vs. qSnow) and recall@1 of 0.962 versus 0.910 in Norland.

As shown in Table \ref{recall}, our CAE is effective and memory-efficient on all datasets, and outperforms NetVLAD and their backbones in most tests. Furthermore, at the dimension of 4096, the dimension of our descriptor is two orders of magnitude smaller than VGG16's 614,400 and AlexNet's 272,384.

\begin{figure*}[ht]
	\includegraphics[width=\textwidth]{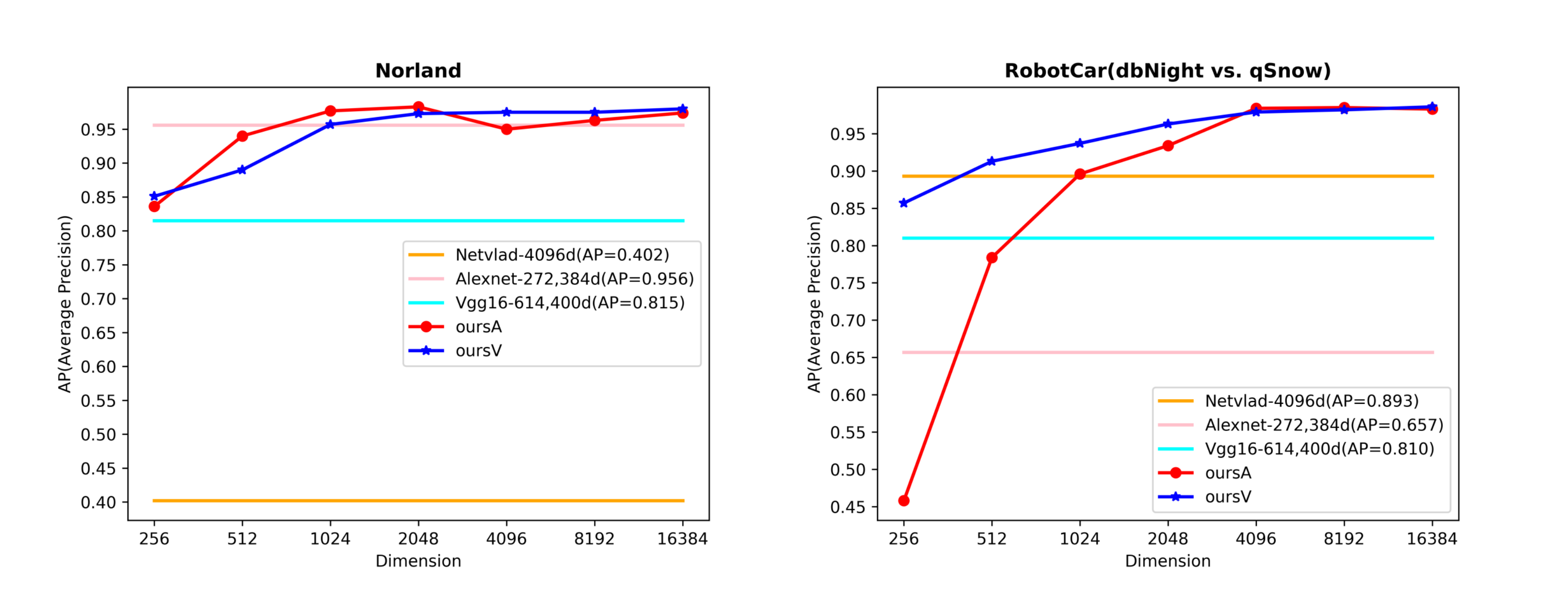}
	\caption{Comparison of the baselines and our methods with different dimensions of our image descriptor in terms of AP on the Norland and RobotCar (dbNight vs. qSnow). \textbf{NetVLAD-4096d(AP=0.402)} means that NetVLAD with 4096 dimensions output can achieve AP of 0.402, and others are similar indications. In the Norland of the left picture, our method can achieve an AP of 0.95 with 1024d. In the RobotCar (dbNight vs. qSnow) of the right picture, OursV with just 1024d can approximately attain an AP of 0.90 as well as NetVLAD. While the dimension of NetVLAD is 4096.}
	\label{dimension}
\end{figure*}

\subsection{Comparison of Encoded Dimensions}

In this section, we will present the results from our study of the relationship between the output dimension of our CAE and matching performance. Fig. \ref{dimension} shows the AP results in different datasets with the variation of the encoded dimensions.
As shown in the left sub-figure, OursV and OursA achieve similar results to AlexNet. However, the output dimension of AlexNet is 272,384. Although the performance of both methods is a bit worse than AlexNet when the dimensions are small, e.g., 512 or 256, they still achieve better results than NetVLAD and VGG16 with a moderate dimension.

From the right sub-figure, we can observe that, even in the urban-scale RobotCar dataset (dbNight vs. qSnow), OursV and OursA can achieve the same results as NetVLAD when the dimension is higher than 1024.
From the above observation, we can infer that our CAE can attain a high performance with low dimensions.
However, as we continue to reduce the descriptor dimension, the performance will will deteriorate.

\begin{figure}[ht]
	\centering
	\subfigure[Norland]{
		\centering
		\includegraphics[width=0.85\textwidth]{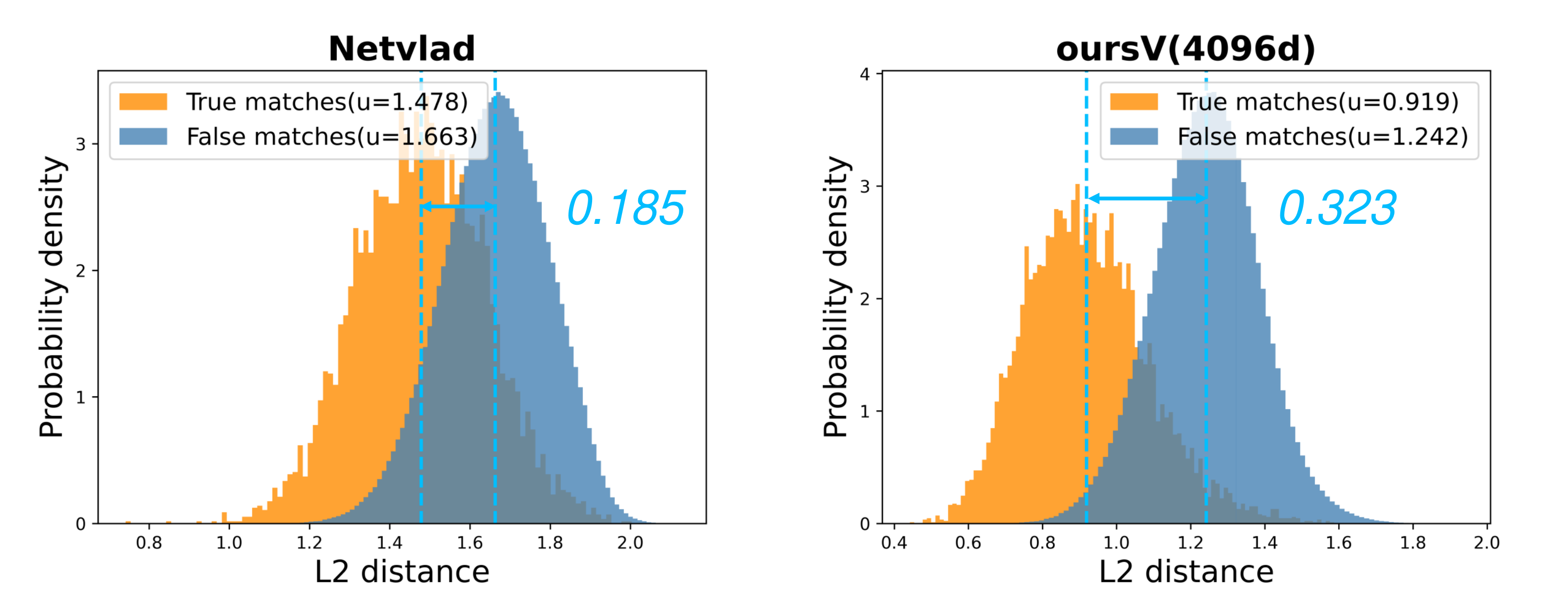}
		\label{l2Distribution-a}
	}
	\subfigure[Robotcar(dbNight vs. qSnow)]{
		\centering
		\includegraphics[width=0.85\textwidth]{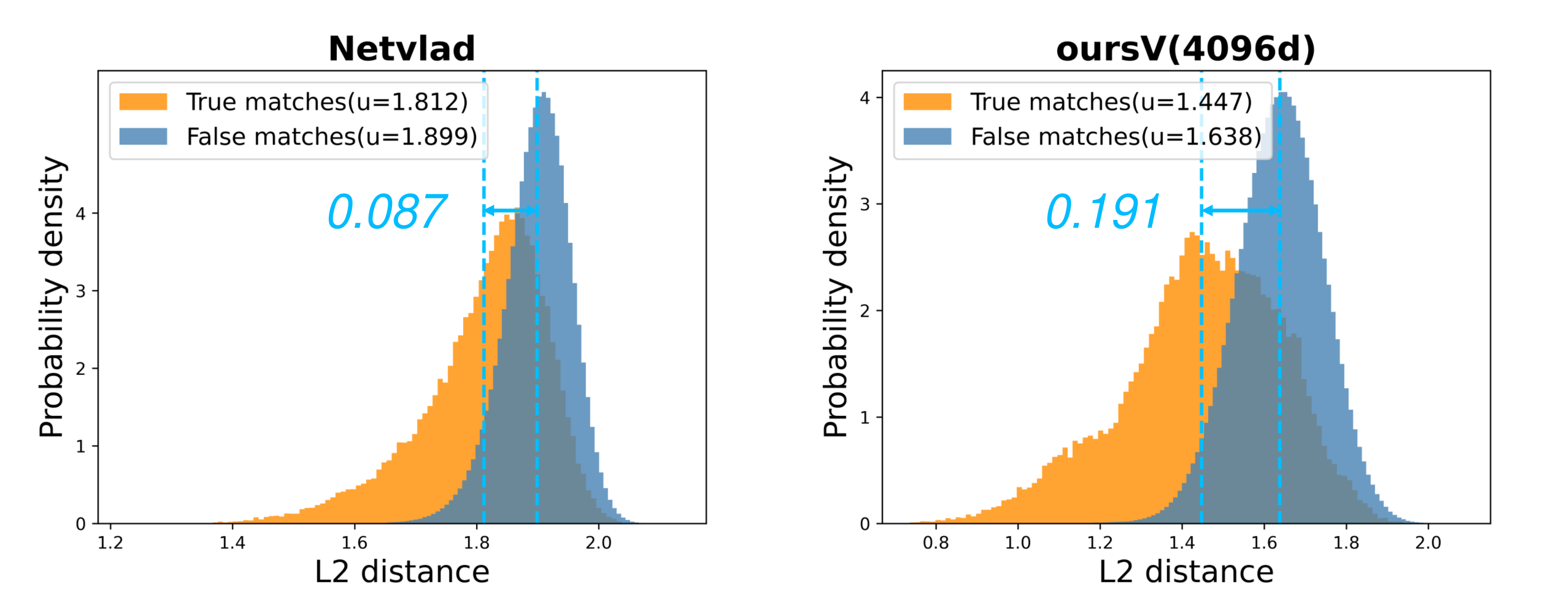}
		\label{l2Distribution-b}
	}
	\caption{L2 distribution of true and false matches for different methods. In both datasets, OursV (4096d) surpass NetVLAD a lot with difference of mean value of 0.323 versus 0.185 in Norland and 0.191 versus 0.087 in RobotCar (dbNight vs. qSnow).}
	\label{l2Distribution}
\end{figure}

\subsection{Discriminative Capacity}
We also plot the distribution of L2-distances between true matches, false matches, to evaluate the discriminating power of our CAE. For a fair comparison, we set the dimension of OursV as 4096, the same as NetVLAD. From Fig. \ref{l2Distribution-a}, we can observe that the overlapping area of OursV is smaller than that of NetVLAD with a mean gap value of 0.323 versus 0.185. As shown in Fig. \ref{l2Distribution-b}, the distributions of L2-distances between the true matches and false matches of NetVLAD are close where half of the true matches overlap with the false matches, resulting in a low mean gap value of 0.087. For OursV, the gap is 0.191, and half of the true matches do not overlap with the false ones.

These results show that our CAE is more discriminative than NetVLAD. However, the distributions of L2-distances between the true and false matches still overlap considerably. This could be caused by the fact that Norland consists of only train road views, while RobotCar is more complicated with dynamic objects.

\begin{figure*}[ht]
	\centering
	\subfigure[Norland]{
		\includegraphics[width=0.45\textwidth]{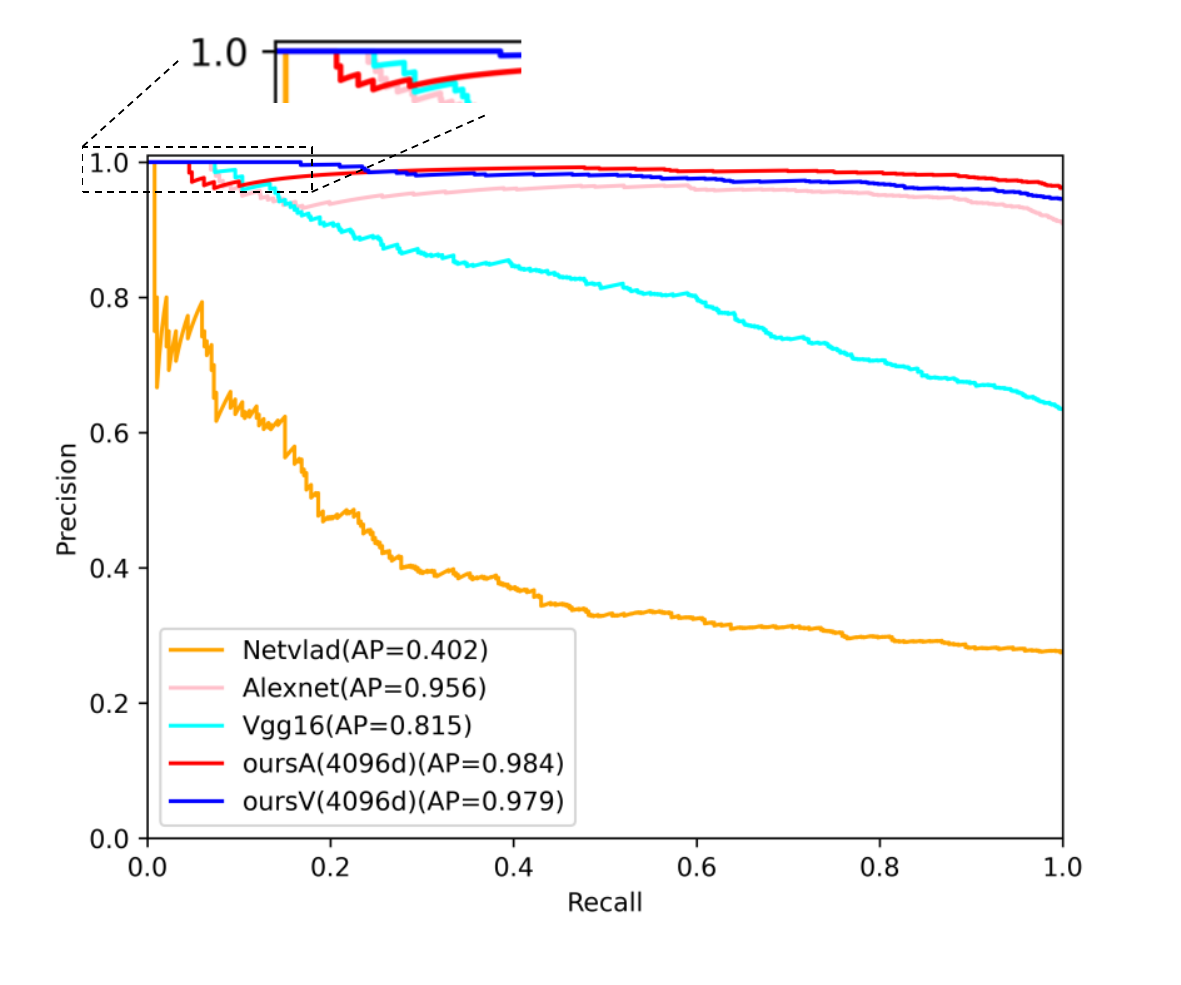}
		\label{pr-curve-a}
	}
	\subfigure[Robotcar(dbNight vs. qSnow)]{
		\includegraphics[width=0.45\textwidth]{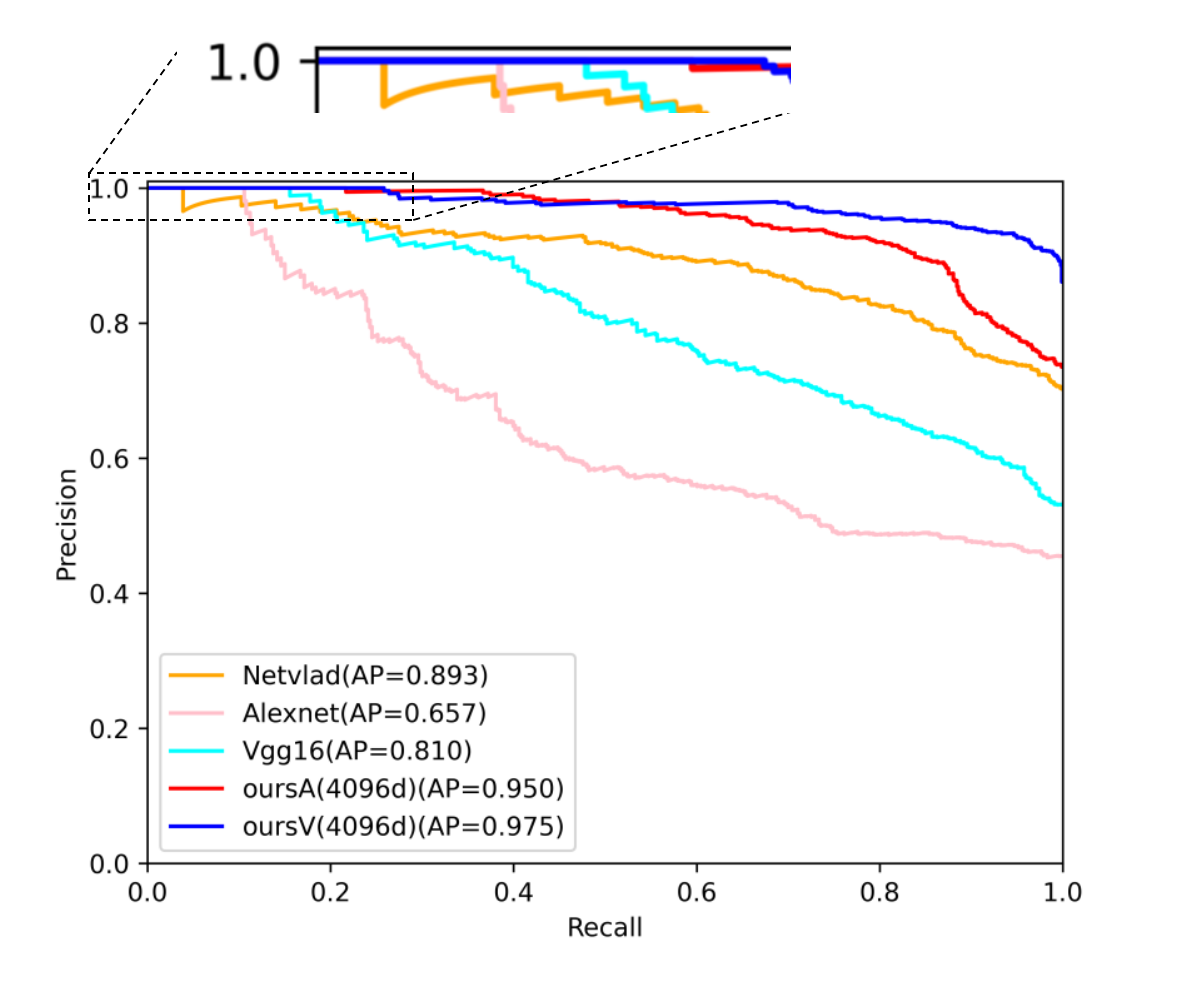}
		\label{pr-curve-b}
	}
	\caption{Precision-Recall Curves for Norland and Robotcar (dbNight vs. qSnow) datasets. The proposed method consistently outperforms better than the baselines with the metric of recall at 100\% precision. In Norland, OursV is the best with almost 0.2 recall at 100\% precision. In Robotcar (dbNight vs. qSnow), OursV and OursA attain nearly 0.3 recall at 100\% precision.}
	\label{pr-curve}
\end{figure*}

\subsection{Ability of False Positives Avoidance}
As mentioned in \cite{lowry2015visual}, false positive matches are fatal to VPR, since false matches lead to incorrect input to robot pose trajectory optimization. Consequently, recall at 100\% precision is the prime metric for many tasks. From the result of the Norland dataset shown in Fig. \ref{pr-curve-a}, OursV surpasses VGG16 and AlexNet in terms of recall at 100\% precision, while OursA and NetVLAD perform poorly in this test. However, as shown in Fig. \ref{pr-curve-b} of a RobotCar (dbNight vs. qSnow) experiment, OursA and OursV perform significantly better than other baselines.

\section{CONCLUSION}
In this paper, we propose a simple method that uses a CAE in constructing an image descriptor from image feature maps from by a CNN.
The experimental results have shown that the compressed CNN-descriptor by the CAE can attain high performance, better than state-of-the-art image descriptors such as NetVLAD and than CNN-based descriptors at much higher dimensions such as VGG16 and AlexNet.
Specifically, our CAE can consistently achieve a higher AP and recall than NetVLAD, when using the same descriptor dimension; in addition, our CAE achieves comparable results to other baseline descriptors when using a lower dimension than these descriptors.
In RobotCar (dbNight vs. qSnow), OursV can achieve top-1 recall of 0.861 with 4096 dimensions, outperforming NetVLAD. Furthermore, from the system perspective, our CAE can achieve higher recall at 100\% precision than others. These quantitative results indicate that dimension reduction by our CAE can produce a compact and condition-invariant global descriptor while reducing the computational cost.

\section*{DECLARATION}
\begin{itemize}
	\item This work was supported in part by the Leading Talents Program of Guangdong Province under Grant No. 2016LJ06G498 and 2019QN01X761
	\item No Conflict of interest/Competing interests (check journal-specific guidelines for which heading to use)
	\item Ethics approval
	\item Consent to participate
	\item Consent for publication
	\item No availability of data and materials
	\item No code availability
	\item Hanjing Ye raised the main idea and completed the experiments and the draft. Weinan Chen and Jingwen Yu provided help with code-work. Li He, Yisheng Guan and Hong Zhang shared their suggestions for revising the idea in this paper.
\end{itemize}

\begin{thebibliography}{35}
		\ifx \bisbn   \undefined \def \bisbn  #1{ISBN #1}\fi
		\ifx \binits  \undefined \def \binits#1{#1}\fi
		\ifx \bauthor  \undefined \def \bauthor#1{#1}\fi
		\ifx \batitle  \undefined \def \batitle#1{#1}\fi
		\ifx \bjtitle  \undefined \def \bjtitle#1{#1}\fi
		\ifx \bvolume  \undefined \def \bvolume#1{\textbf{#1}}\fi
		\ifx \byear  \undefined \def \byear#1{#1}\fi
		\ifx \bissue  \undefined \def \bissue#1{#1}\fi
		\ifx \bfpage  \undefined \def \bfpage#1{#1}\fi
		\ifx \blpage  \undefined \def \blpage #1{#1}\fi
		\ifx \burl  \undefined \def \burl#1{\textsf{#1}}\fi
		\ifx \doiurl  \undefined \def \doiurl#1{\url{https://doi.org/#1}}\fi
		\ifx \betal  \undefined \def \betal{\textit{et al.}}\fi
		\ifx \binstitute  \undefined \def \binstitute#1{#1}\fi
		\ifx \binstitutionaled  \undefined \def \binstitutionaled#1{#1}\fi
		\ifx \bctitle  \undefined \def \bctitle#1{#1}\fi
		\ifx \beditor  \undefined \def \beditor#1{#1}\fi
		\ifx \bpublisher  \undefined \def \bpublisher#1{#1}\fi
		\ifx \bbtitle  \undefined \def \bbtitle#1{#1}\fi
		\ifx \bedition  \undefined \def \bedition#1{#1}\fi
		\ifx \bseriesno  \undefined \def \bseriesno#1{#1}\fi
		\ifx \blocation  \undefined \def \blocation#1{#1}\fi
		\ifx \bsertitle  \undefined \def \bsertitle#1{#1}\fi
		\ifx \bsnm \undefined \def \bsnm#1{#1}\fi
		\ifx \bsuffix \undefined \def \bsuffix#1{#1}\fi
		\ifx \bparticle \undefined \def \bparticle#1{#1}\fi
		\ifx \barticle \undefined \def \barticle#1{#1}\fi
		\bibcommenthead
		\ifx \bconfdate \undefined \def \bconfdate #1{#1}\fi
		\ifx \botherref \undefined \def \botherref #1{#1}\fi
		\ifx \url \undefined \def \url#1{\textsf{#1}}\fi
		\ifx \bchapter \undefined \def \bchapter#1{#1}\fi
		\ifx \bbook \undefined \def \bbook#1{#1}\fi
		\ifx \bcomment \undefined \def \bcomment#1{#1}\fi
		\ifx \oauthor \undefined \def \oauthor#1{#1}\fi
		\ifx \citeauthoryear \undefined \def \citeauthoryear#1{#1}\fi
		\ifx \endbibitem  \undefined \def \endbibitem {}\fi
		\ifx \bconflocation  \undefined \def \bconflocation#1{#1}\fi
		\ifx \arxivurl  \undefined \def \arxivurl#1{\textsf{#1}}\fi
		\csname PreBibitemsHook\endcsname
		
		\bibitem{olid2018single}
		\begin{botherref}
			\oauthor{\bsnm{Olid}, \binits{D.}},
			\oauthor{\bsnm{F{\'a}cil}, \binits{J.M.}},
			\oauthor{\bsnm{Civera}, \binits{J.}}:
			Single-view place recognition under seasonal changes.
			arXiv preprint arXiv:1808.06516
			(2018)
		\end{botherref}
		\endbibitem
		
		\bibitem{liu2015keypoint}
		\begin{bchapter}
			\bauthor{\bsnm{Liu}, \binits{Y.}},
			\bauthor{\bsnm{Feng}, \binits{R.}},
			\bauthor{\bsnm{Zhang}, \binits{H.}}:
			\bctitle{Keypoint matching by outlier pruning with consensus constraint}.
			In: \bbtitle{2015 IEEE International Conference on Robotics and Automation
				(ICRA)},
			pp. \bfpage{5481}--\blpage{5486}
			(\byear{2015}).
			\bcomment{IEEE}
		\end{bchapter}
		\endbibitem
		
		\bibitem{maddern20171}
		\begin{barticle}
			\bauthor{\bsnm{Maddern}, \binits{W.}},
			\bauthor{\bsnm{Pascoe}, \binits{G.}},
			\bauthor{\bsnm{Linegar}, \binits{C.}},
			\bauthor{\bsnm{Newman}, \binits{P.}}:
			\batitle{1 year, 1000 km: The oxford robotcar dataset}.
			\bjtitle{The International Journal of Robotics Research}
			\bvolume{36}(\bissue{1}),
			\bfpage{3}--\blpage{15}
			(\byear{2017})
		\end{barticle}
		\endbibitem
		
		\bibitem{lowry2015visual}
		\begin{barticle}
			\bauthor{\bsnm{Lowry}, \binits{S.}},
			\bauthor{\bsnm{S{\"u}nderhauf}, \binits{N.}},
			\bauthor{\bsnm{Newman}, \binits{P.}},
			\bauthor{\bsnm{Leonard}, \binits{J.J.}},
			\bauthor{\bsnm{Cox}, \binits{D.}},
			\bauthor{\bsnm{Corke}, \binits{P.}},
			\bauthor{\bsnm{Milford}, \binits{M.J.}}:
			\batitle{Visual place recognition: A survey}.
			\bjtitle{IEEE Transactions on Robotics}
			\bvolume{32}(\bissue{1}),
			\bfpage{1}--\blpage{19}
			(\byear{2015})
		\end{barticle}
		\endbibitem
		
		\bibitem{lowe2004distinctive}
		\begin{barticle}
			\bauthor{\bsnm{Lowe}, \binits{D.G.}}:
			\batitle{Distinctive image features from scale-invariant keypoints}.
			\bjtitle{International journal of computer vision}
			\bvolume{60}(\bissue{2}),
			\bfpage{91}--\blpage{110}
			(\byear{2004})
		\end{barticle}
		\endbibitem
		
		\bibitem{alahi2012freak}
		\begin{bchapter}
			\bauthor{\bsnm{Alahi}, \binits{A.}},
			\bauthor{\bsnm{Ortiz}, \binits{R.}},
			\bauthor{\bsnm{Vandergheynst}, \binits{P.}}:
			\bctitle{Freak: Fast retina keypoint}.
			In: \bbtitle{2012 IEEE Conference on Computer Vision and Pattern Recognition},
			pp. \bfpage{510}--\blpage{517}
			(\byear{2012}).
			\bcomment{Ieee}
		\end{bchapter}
		\endbibitem
		
		\bibitem{rublee2011orb}
		\begin{bchapter}
			\bauthor{\bsnm{Rublee}, \binits{E.}},
			\bauthor{\bsnm{Rabaud}, \binits{V.}},
			\bauthor{\bsnm{Konolige}, \binits{K.}},
			\bauthor{\bsnm{Bradski}, \binits{G.}}:
			\bctitle{Orb: An efficient alternative to sift or surf}.
			In: \bbtitle{2011 International Conference on Computer Vision},
			pp. \bfpage{2564}--\blpage{2571}
			(\byear{2011}).
			\bcomment{Ieee}
		\end{bchapter}
		\endbibitem
		
		\bibitem{bay2008speeded}
		\begin{barticle}
			\bauthor{\bsnm{Bay}, \binits{H.}},
			\bauthor{\bsnm{Ess}, \binits{A.}},
			\bauthor{\bsnm{Tuytelaars}, \binits{T.}},
			\bauthor{\bsnm{Van~Gool}, \binits{L.}}:
			\batitle{Speeded-up robust features (surf)}.
			\bjtitle{Computer vision and image understanding}
			\bvolume{110}(\bissue{3}),
			\bfpage{346}--\blpage{359}
			(\byear{2008})
		\end{barticle}
		\endbibitem
		
		\bibitem{krizhevsky2012imagenet}
		\begin{barticle}
			\bauthor{\bsnm{Krizhevsky}, \binits{A.}},
			\bauthor{\bsnm{Sutskever}, \binits{I.}},
			\bauthor{\bsnm{Hinton}, \binits{G.E.}}:
			\batitle{Imagenet classification with deep convolutional neural networks}.
			\bjtitle{Advances in neural information processing systems}
			\bvolume{25},
			\bfpage{1097}--\blpage{1105}
			(\byear{2012})
		\end{barticle}
		\endbibitem
		
		\bibitem{girshick2014rich}
		\begin{bchapter}
			\bauthor{\bsnm{Girshick}, \binits{R.}},
			\bauthor{\bsnm{Donahue}, \binits{J.}},
			\bauthor{\bsnm{Darrell}, \binits{T.}},
			\bauthor{\bsnm{Malik}, \binits{J.}}:
			\bctitle{Rich feature hierarchies for accurate object detection and semantic
				segmentation}.
			In: \bbtitle{Proceedings of the IEEE Conference on Computer Vision and Pattern
				Recognition},
			pp. \bfpage{580}--\blpage{587}
			(\byear{2014})
		\end{bchapter}
		\endbibitem
		
		\bibitem{ronneberger2015u}
		\begin{bchapter}
			\bauthor{\bsnm{Ronneberger}, \binits{O.}},
			\bauthor{\bsnm{Fischer}, \binits{P.}},
			\bauthor{\bsnm{Brox}, \binits{T.}}:
			\bctitle{U-net: Convolutional networks for biomedical image segmentation}.
			In: \bbtitle{International Conference on Medical Image Computing and
				Computer-assisted Intervention},
			pp. \bfpage{234}--\blpage{241}
			(\byear{2015}).
			\bcomment{Springer}
		\end{bchapter}
		\endbibitem
		
		\bibitem{sunderhauf2015performance}
		\begin{bchapter}
			\bauthor{\bsnm{S{\"u}nderhauf}, \binits{N.}},
			\bauthor{\bsnm{Shirazi}, \binits{S.}},
			\bauthor{\bsnm{Dayoub}, \binits{F.}},
			\bauthor{\bsnm{Upcroft}, \binits{B.}},
			\bauthor{\bsnm{Milford}, \binits{M.}}:
			\bctitle{On the performance of convnet features for place recognition}.
			In: \bbtitle{2015 IEEE/RSJ International Conference on Intelligent Robots and
				Systems (IROS)},
			pp. \bfpage{4297}--\blpage{4304}
			(\byear{2015}).
			\bcomment{IEEE}
		\end{bchapter}
		\endbibitem
		
		\bibitem{arandjelovic2016NetVLAD}
		\begin{bchapter}
			\bauthor{\bsnm{Arandjelovic}, \binits{R.}},
			\bauthor{\bsnm{Gronat}, \binits{P.}},
			\bauthor{\bsnm{Torii}, \binits{A.}},
			\bauthor{\bsnm{Pajdla}, \binits{T.}},
			\bauthor{\bsnm{Sivic}, \binits{J.}}:
			\bctitle{Netvlad: Cnn architecture for weakly supervised place recognition}.
			In: \bbtitle{Proceedings of the IEEE Conference on Computer Vision and Pattern
				Recognition},
			pp. \bfpage{5297}--\blpage{5307}
			(\byear{2016})
		\end{bchapter}
		\endbibitem
		
		\bibitem{radenovic2018fine}
		\begin{barticle}
			\bauthor{\bsnm{Radenovi{\'c}}, \binits{F.}},
			\bauthor{\bsnm{Tolias}, \binits{G.}},
			\bauthor{\bsnm{Chum}, \binits{O.}}:
			\batitle{Fine-tuning cnn image retrieval with no human annotation}.
			\bjtitle{IEEE transactions on pattern analysis and machine intelligence}
			\bvolume{41}(\bissue{7}),
			\bfpage{1655}--\blpage{1668}
			(\byear{2018})
		\end{barticle}
		\endbibitem
		
		\bibitem{gordo2016deep}
		\begin{bchapter}
			\bauthor{\bsnm{Gordo}, \binits{A.}},
			\bauthor{\bsnm{Almaz{\'a}n}, \binits{J.}},
			\bauthor{\bsnm{Revaud}, \binits{J.}},
			\bauthor{\bsnm{Larlus}, \binits{D.}}:
			\bctitle{Deep image retrieval: Learning global representations for image
				search}.
			In: \bbtitle{European Conference on Computer Vision},
			pp. \bfpage{241}--\blpage{257}
			(\byear{2016}).
			\bcomment{Springer}
		\end{bchapter}
		\endbibitem
		
		\bibitem{sivic2003video}
		\begin{bchapter}
			\bauthor{\bsnm{Sivic}, \binits{J.}},
			\bauthor{\bsnm{Zisserman}, \binits{A.}}:
			\bctitle{Video google: A text retrieval approach to object matching in videos}.
			In: \bbtitle{Computer Vision, IEEE International Conference On},
			vol. \bseriesno{3},
			pp. \bfpage{1470}--\blpage{1470}
			(\byear{2003}).
			\bcomment{IEEE Computer Society}
		\end{bchapter}
		\endbibitem
		
		\bibitem{jegou2010aggregating}
		\begin{bchapter}
			\bauthor{\bsnm{J{\'e}gou}, \binits{H.}},
			\bauthor{\bsnm{Douze}, \binits{M.}},
			\bauthor{\bsnm{Schmid}, \binits{C.}},
			\bauthor{\bsnm{P{\'e}rez}, \binits{P.}}:
			\bctitle{Aggregating local descriptors into a compact image representation}.
			In: \bbtitle{2010 IEEE Computer Society Conference on Computer Vision and
				Pattern Recognition},
			pp. \bfpage{3304}--\blpage{3311}
			(\byear{2010}).
			\bcomment{IEEE}
		\end{bchapter}
		\endbibitem
		
		\bibitem{philbin2007object}
		\begin{bchapter}
			\bauthor{\bsnm{Philbin}, \binits{J.}},
			\bauthor{\bsnm{Chum}, \binits{O.}},
			\bauthor{\bsnm{Isard}, \binits{M.}},
			\bauthor{\bsnm{Sivic}, \binits{J.}},
			\bauthor{\bsnm{Zisserman}, \binits{A.}}:
			\bctitle{Object retrieval with large vocabularies and fast spatial matching}.
			In: \bbtitle{2007 IEEE Conference on Computer Vision and Pattern Recognition},
			pp. \bfpage{1}--\blpage{8}
			(\byear{2007}).
			\bcomment{IEEE}
		\end{bchapter}
		\endbibitem
		
		\bibitem{arandjelovic2013all}
		\begin{bchapter}
			\bauthor{\bsnm{Arandjelovic}, \binits{R.}},
			\bauthor{\bsnm{Zisserman}, \binits{A.}}:
			\bctitle{All about vlad}.
			In: \bbtitle{Proceedings of the IEEE Conference on Computer Vision and Pattern
				Recognition},
			pp. \bfpage{1578}--\blpage{1585}
			(\byear{2013})
		\end{bchapter}
		\endbibitem
		
		\bibitem{jegou2011aggregating}
		\begin{barticle}
			\bauthor{\bsnm{J{\'e}gou}, \binits{H.}},
			\bauthor{\bsnm{Perronnin}, \binits{F.}},
			\bauthor{\bsnm{Douze}, \binits{M.}},
			\bauthor{\bsnm{S{\'a}nchez}, \binits{J.}},
			\bauthor{\bsnm{P{\'e}rez}, \binits{P.}},
			\bauthor{\bsnm{Schmid}, \binits{C.}}:
			\batitle{Aggregating local image descriptors into compact codes}.
			\bjtitle{IEEE transactions on pattern analysis and machine intelligence}
			\bvolume{34}(\bissue{9}),
			\bfpage{1704}--\blpage{1716}
			(\byear{2011})
		\end{barticle}
		\endbibitem
		
		\bibitem{sharif2014cnn}
		\begin{bchapter}
			\bauthor{\bsnm{Sharif~Razavian}, \binits{A.}},
			\bauthor{\bsnm{Azizpour}, \binits{H.}},
			\bauthor{\bsnm{Sullivan}, \binits{J.}},
			\bauthor{\bsnm{Carlsson}, \binits{S.}}:
			\bctitle{Cnn features off-the-shelf: an astounding baseline for recognition}.
			In: \bbtitle{Proceedings of the IEEE Conference on Computer Vision and Pattern
				Recognition Workshops},
			pp. \bfpage{806}--\blpage{813}
			(\byear{2014})
		\end{bchapter}
		\endbibitem
		
		\bibitem{babenko2014neural}
		\begin{bchapter}
			\bauthor{\bsnm{Babenko}, \binits{A.}},
			\bauthor{\bsnm{Slesarev}, \binits{A.}},
			\bauthor{\bsnm{Chigorin}, \binits{A.}},
			\bauthor{\bsnm{Lempitsky}, \binits{V.}}:
			\bctitle{Neural codes for image retrieval}.
			In: \bbtitle{European Conference on Computer Vision},
			pp. \bfpage{584}--\blpage{599}
			(\byear{2014}).
			\bcomment{Springer}
		\end{bchapter}
		\endbibitem
		
		\bibitem{tolias2015particular}
		\begin{botherref}
			\oauthor{\bsnm{Tolias}, \binits{G.}},
			\oauthor{\bsnm{Sicre}, \binits{R.}},
			\oauthor{\bsnm{J{\'e}gou}, \binits{H.}}:
			Particular object retrieval with integral max-pooling of cnn activations.
			arXiv preprint arXiv:1511.05879
			(2015)
		\end{botherref}
		\endbibitem
		
		\bibitem{razavian2016visual}
		\begin{barticle}
			\bauthor{\bsnm{Razavian}, \binits{A.S.}},
			\bauthor{\bsnm{Sullivan}, \binits{J.}},
			\bauthor{\bsnm{Carlsson}, \binits{S.}},
			\bauthor{\bsnm{Maki}, \binits{A.}}:
			\batitle{Visual instance retrieval with deep convolutional networks}.
			\bjtitle{ITE Transactions on Media Technology and Applications}
			\bvolume{4}(\bissue{3}),
			\bfpage{251}--\blpage{258}
			(\byear{2016})
		\end{barticle}
		\endbibitem
		
		\bibitem{mirza2014conditional}
		\begin{botherref}
			\oauthor{\bsnm{Mirza}, \binits{M.}},
			\oauthor{\bsnm{Osindero}, \binits{S.}}:
			Conditional generative adversarial nets.
			arXiv preprint arXiv:1411.1784
			(2014)
		\end{botherref}
		\endbibitem
		
		\bibitem{isola2017image}
		\begin{bchapter}
			\bauthor{\bsnm{Isola}, \binits{P.}},
			\bauthor{\bsnm{Zhu}, \binits{J.-Y.}},
			\bauthor{\bsnm{Zhou}, \binits{T.}},
			\bauthor{\bsnm{Efros}, \binits{A.A.}}:
			\bctitle{Image-to-image translation with conditional adversarial networks}.
			In: \bbtitle{Proceedings of the IEEE Conference on Computer Vision and Pattern
				Recognition},
			pp. \bfpage{1125}--\blpage{1134}
			(\byear{2017})
		\end{bchapter}
		\endbibitem
		
		\bibitem{vankadari2020unsupervised}
		\begin{bchapter}
			\bauthor{\bsnm{Vankadari}, \binits{M.}},
			\bauthor{\bsnm{Garg}, \binits{S.}},
			\bauthor{\bsnm{Majumder}, \binits{A.}},
			\bauthor{\bsnm{Kumar}, \binits{S.}},
			\bauthor{\bsnm{Behera}, \binits{A.}}:
			\bctitle{Unsupervised monocular depth estimation for night-time images using
				adversarial domain feature adaptation}.
			In: \bbtitle{European Conference on Computer Vision},
			pp. \bfpage{443}--\blpage{459}
			(\byear{2020}).
			\bcomment{Springer}
		\end{bchapter}
		\endbibitem
		
		\bibitem{merrill2018lightweight}
		\begin{botherref}
			\oauthor{\bsnm{Merrill}, \binits{N.}},
			\oauthor{\bsnm{Huang}, \binits{G.}}:
			Lightweight unsupervised deep loop closure.
			arXiv preprint arXiv:1805.07703
			(2018)
		\end{botherref}
		\endbibitem
		
		\bibitem{dai2020keypoint}
		\begin{bchapter}
			\bauthor{\bsnm{Dai}, \binits{Z.}},
			\bauthor{\bsnm{Huang}, \binits{X.}},
			\bauthor{\bsnm{Chen}, \binits{W.}},
			\bauthor{\bsnm{Chen}, \binits{C.}},
			\bauthor{\bsnm{He}, \binits{L.}},
			\bauthor{\bsnm{Wen}, \binits{S.}},
			\bauthor{\bsnm{Zhang}, \binits{H.}}:
			\bctitle{Keypoint description by descriptor fusion using autoencoders}.
			In: \bbtitle{2020 IEEE International Conference on Robotics and Automation
				(ICRA)},
			pp. \bfpage{65}--\blpage{71}
			(\byear{2020}).
			\bcomment{IEEE}
		\end{bchapter}
		\endbibitem
		
		\bibitem{ba2016layer}
		\begin{botherref}
			\oauthor{\bsnm{Ba}, \binits{J.L.}},
			\oauthor{\bsnm{Kiros}, \binits{J.R.}},
			\oauthor{\bsnm{Hinton}, \binits{G.E.}}:
			Layer normalization.
			arXiv preprint arXiv:1607.06450
			(2016)
		\end{botherref}
		\endbibitem
		
		\bibitem{hou2015convolutional}
		\begin{bchapter}
			\bauthor{\bsnm{Hou}, \binits{Y.}},
			\bauthor{\bsnm{Zhang}, \binits{H.}},
			\bauthor{\bsnm{Zhou}, \binits{S.}}:
			\bctitle{Convolutional neural network-based image representation for visual
				loop closure detection}.
			In: \bbtitle{2015 IEEE International Conference on Information and Automation},
			pp. \bfpage{2238}--\blpage{2245}
			(\byear{2015}).
			\bcomment{IEEE}
		\end{bchapter}
		\endbibitem
		
		\bibitem{simonyan2014very}
		\begin{botherref}
			\oauthor{\bsnm{Simonyan}, \binits{K.}},
			\oauthor{\bsnm{Zisserman}, \binits{A.}}:
			Very deep convolutional networks for large-scale image recognition.
			arXiv preprint arXiv:1409.1556
			(2014)
		\end{botherref}
		\endbibitem
		
		\bibitem{chenonly2017}
		\begin{botherref}
			\oauthor{\bsnm{Chen}, \binits{Z.}},
			\oauthor{\bsnm{Maffra}, \binits{F.}},
			\oauthor{\bsnm{Sa}, \binits{I.}},
			\oauthor{\bsnm{Chli}, \binits{M.}}:
			Only look once, mining distinctive landmarks from convnet for visual place
			recognition. in 2017 ieee.
			In: RSJ International Conference on Intelligent Robots and Systems (IROS),
			pp. 9--16
		\end{botherref}
		\endbibitem
		
		\bibitem{ioffe2015batch}
		\begin{bchapter}
			\bauthor{\bsnm{Ioffe}, \binits{S.}},
			\bauthor{\bsnm{Szegedy}, \binits{C.}}:
			\bctitle{Batch normalization: Accelerating deep network training by reducing
				internal covariate shift}.
			In: \bbtitle{International Conference on Machine Learning},
			pp. \bfpage{448}--\blpage{456}
			(\byear{2015}).
			\bcomment{PMLR}
		\end{bchapter}
		\endbibitem
		
		\bibitem{he2015delving}
		\begin{bchapter}
			\bauthor{\bsnm{He}, \binits{K.}},
			\bauthor{\bsnm{Zhang}, \binits{X.}},
			\bauthor{\bsnm{Ren}, \binits{S.}},
			\bauthor{\bsnm{Sun}, \binits{J.}}:
			\bctitle{Delving deep into rectifiers: Surpassing human-level performance on
				imagenet classification}.
			In: \bbtitle{Proceedings of the IEEE International Conference on Computer
				Vision},
			pp. \bfpage{1026}--\blpage{1034}
			(\byear{2015})
		\end{bchapter}
		\endbibitem
		
\end{thebibliography}



\end{document}